\documentclass[letterpaper, 10 pt, conference]{ieeeconf}  

\IEEEoverridecommandlockouts                              

\usepackage[dvipsnames]{xcolor}
\usepackage{amsmath}
\usepackage{amssymb}  
\usepackage{bbm}  
\usepackage{graphicx}
\usepackage{tikz}
\usepackage{pgfplots}
\usepackage{csvsimple}
\usepackage{color}
\usepackage{subfig}
\usepackage{url}
\usepackage{mathtools}
\usepackage{balance}
\usepackage[hidelinks]{hyperref}

\usetikzlibrary{calc, fit}
\usepackage{tikz-3dplot}
\usetikzlibrary{positioning, 3d}
\usetikzlibrary{backgrounds}

\newcommand\mat[1]{\mathbf{#1}}

\newcommand{\etal}{{\em et al.}}

\DeclareMathOperator*{\minimize}{minimize}

\definecolor{aisblue}{RGB}{51,102,153}
\definecolor{aisred}{RGB}{154,0,0}

\definecolor{color1}{HTML}{4D4D4D} 
\definecolor{color2}{HTML}{F15854} 
\definecolor{color3}{HTML}{5DA5DA} 
\definecolor{color4}{HTML}{FAA43A} 
\definecolor{color5}{HTML}{60BD68} 
\definecolor{color6}{HTML}{B2912F} 
\definecolor{color7}{HTML}{B276B2} 
\definecolor{color8}{HTML}{DECF3F} 
\definecolor{color9}{HTML}{F17CB0} 

\title{\LARGE \bf
Optimization Beyond the Convolution: Generalizing Spatial Relations with End-to-End Metric Learning
}

\author{Philipp Jund, Andreas Eitel, Nichola Abdo and Wolfram Burgard$^{1}$%
\thanks{This work has been supported by the German Research Foundation under research unit FOR 1513 (HYBRIS) and the priority program SPP 1527.}
\thanks{$^{1}$All authors are with the Department of Computer Science, University of Freiburg, Germany.
        {\footnotesize {\{jundp, eitel, abdon, burgard\}@cs.uni-freiburg.de}}}%
}

\begin{document}

\maketitle
\thispagestyle{empty}
\pagestyle{empty}

\begin{abstract}

To operate intelligently in domestic environments, robots require the ability to understand 
arbitrary spatial relations between objects and to generalize them to objects of varying sizes 
and shapes. In this work, we present a novel end-to-end approach to generalize spatial relations 
based on distance metric learning. We train a neural network to transform 3D point clouds of objects 
to a metric space that captures the similarity of the depicted spatial relations, using only geometric 
models of the objects. Our approach employs gradient-based optimization to compute object poses in 
order to imitate an arbitrary target relation by reducing the distance to it under the learned metric. 
Our results based on simulated and real-world experiments show that the proposed method enables robots to 
generalize spatial relations to unknown objects over a continuous spectrum.

\end{abstract}

\section{Introduction}

\label{sec:introduction}
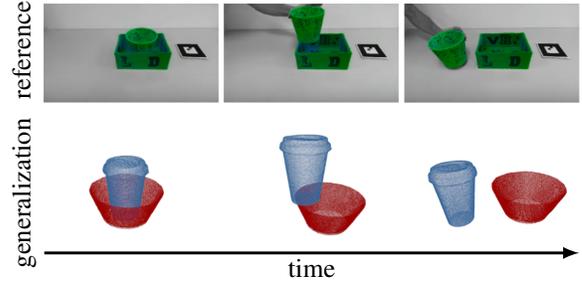
\begin{figure}
\begin{center}
\begin{tikzpicture}

\node (fig) at (0,0) {
            \includegraphics[width=0.4\textwidth, keepaspectratio, interpolate=false]{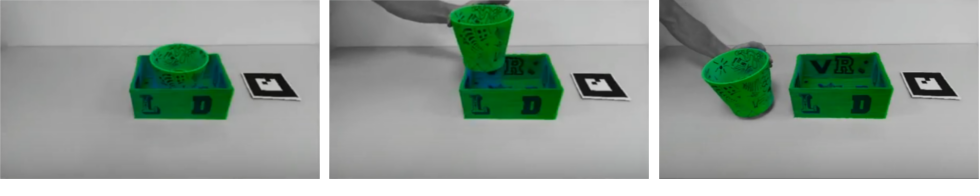}};

\node (fig1) at (0,-1.7) {
            \includegraphics[width=0.4\textwidth, keepaspectratio, interpolate=false]{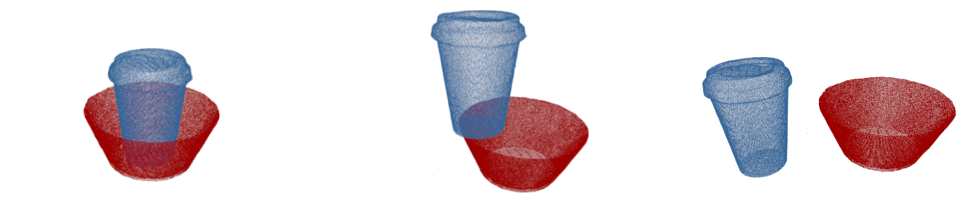}};

\node (a) [left=12pt of fig, yshift=0.1cm, rotate=90, align=left, anchor=north] {reference};
\node (b) [left=12pt of fig1, yshift=-0.1cm, rotate=90, align=left, anchor=north] {generalization};
\draw[->, >=latex, line width=1.1pt] (-0.2\textwidth, -2.6) -- (0.2\textwidth, -2.6);
\node (time) at (0, -2.8) {time};
\end{tikzpicture}

\caption{The goal of our work is to enable a robot to imitate arbitrary spatial relations between pairs of objects and to generalize them to objects of different shapes and sizes. Top: three consecutive, arbitrary relations we presented our approach with, which we perceive using a Kinect2 camera. Bottom: the corresponding generalization of the relations using two new objects as computed by our approach.}
\label{fig:SmallCover}
\end{center}
\end{figure}

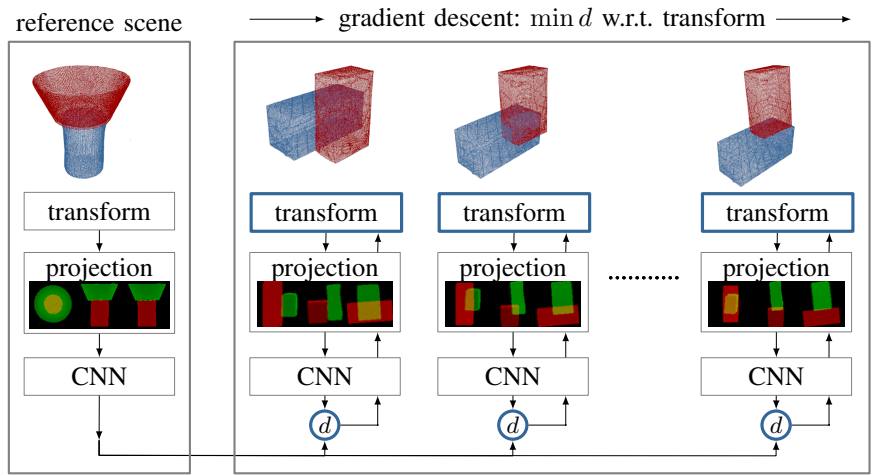
\begin{figure*}[!tp]
    \begin{center}
    \begin{tikzpicture}[line join=round, very thin, every node/.style = {anchor= north west, draw=black!50}, >=latex]

  \node (pc) at (0, 4.6) [draw=none]{\includegraphics[width=1.3cm]{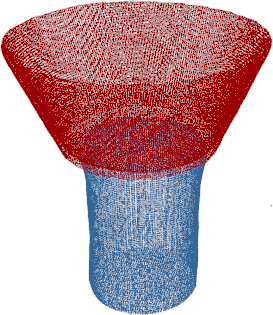}};
  \node (transform) at (0,2.8) [draw=black!50,minimum width=2cm,minimum height=0.5cm] {transform};

  \node (projlayer) at (0,2) [draw,minimum width=2cm,minimum height=0.5cm, align=center, inner sep=2pt] {
          projection \\
          \includegraphics[width=0.62cm, interpolate=false]{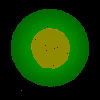}\includegraphics[width=0.62cm, interpolate=false]{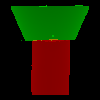}\includegraphics[width=0.62cm, interpolate=false]{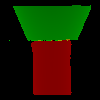}};

  \node (cnn) at (0,0.6) [draw=black!50,minimum width=2cm,minimum height=0.5cm] {CNN};
  \node(dist) at (0 + 1,-0.3) [anchor=center, draw=none, circle, inner sep=1pt] {{\color{white}d}};
  
  \draw[->] (transform) -- (projlayer);
  \draw[->] (projlayer) -- (cnn);
  \draw[->] (cnn) -- (dist.south);

  \newcommand\convnet[6]{
  \def\startx{#2}

  \node (pc#1)[draw=none] at (\startx+0.1, 4.6) {\includegraphics[height=#6]{figures/cover/#5}};
  \node (transform#1) at (\startx,2.8) [minimum width=2cm,minimum height=0.5cm, draw=#4] {transform};

  \node (projlayer#1) at (\startx,2) [draw,minimum width=2cm,minimum height=0.5cm, align=center, inner sep=2pt] {
          projection \\
          \includegraphics[width=0.60cm, interpolate=false]{figures/cover/#3_0.png}\includegraphics[width=0.60cm, interpolate=false]{figures/cover/#3_1.png}\includegraphics[width=0.60cm, interpolate=false]{figures/cover/#3_2.png}};

  \node (cnn#1) at (\startx,0.6) [draw,minimum width=2cm,minimum height=0.5cm] {CNN};
  \node(dist#1) at (\startx + 1,-0.3) [anchor=center, draw, circle, inner sep=1pt, draw=aisblue,very thick] {\small{$d$}};

  \draw[->] (transform#1) -- (projlayer#1);
  \draw[->] (projlayer#1) -- (cnn#1);
  \draw[->] (cnn#1) -- (dist#1);
  \draw[loosely dotted, very thick, line cap=round, dash pattern=on 0pt off 2\pgflinewidth] (7.8, 1.66) -- (8.7, 1.66);
  
  \draw [->] (dist#1.east) |- ([shift={(5mm,0mm)}]dist#1.east) -| ([shift={(7mm,0mm)}]cnn#1.south);

  \draw[->] ([shift={(7mm,0mm)}]projlayer#1.north) -- ([shift={(6.9mm,0mm)}]transform#1.south);
  \draw[->] ([shift={(7mm,0mm)}]cnn#1.north) -- ([shift={(7mm,0mm)}]projlayer#1.south);
  \draw[->] (dist.south) |- ([shift={(0mm,-2mm)}]dist.south) -- ([shift={(0mm,-2mm)}]dist#1.south) -| (dist#1.south);

  }
  \convnet{3}{3}{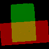}{aisblue, very thick}{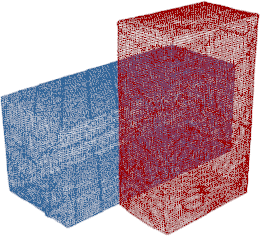}{1.3cm}
  \convnet{5}{5.5}{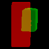}{aisblue, very thick}{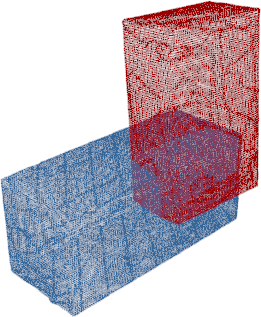}{1.5cm}
  \convnet{9}{9}{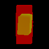}{aisblue, very thick}{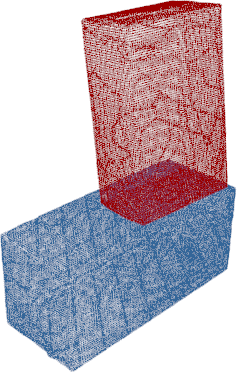}{1.6cm}

  \node (dummy) [draw=none] at (5, -0.5) {};
  \node (dummy2) [draw=none] at (0, -0.5) {};
  \node[draw=black!50, thick, inner sep=2mm,label=above:reference scene,fit=(pc) (dummy2.south) (transform) (transform)] {};
  \node[draw=black!50, thick,inner sep=2mm,label=above:gradient descent: $\min d$ w.r.t. transform,fit=(pc3) (dummy.south) (transform3) (transform9)] {};
  \draw[->] (3, 5.1) -- (4, 5.1);
  \draw[->] (10, 5.1) -- (11, 5.1);

\end{tikzpicture}
    \end{center}
    \caption[Architecture Overview]{An overview of our approach to generalize a relation. The transformation to the metric space consists of a function applying the 3D transformations, a projection of the point clouds to depth images, and a convolutional network pre-trained on pairs of relations.
    During test time, we backpropagate the error of the euclidean distance between the test scene's embeddings and the reference scene's embeddings, to optimize 3D translation and rotation of two objects to resemble the reference scene's spatial relations. Supplementary video: \href{http://spatialrelations.cs.uni-freiburg.de}{\url{http://spatialrelations.cs.uni-freiburg.de}}}
    \label{fig:Cover}
\end{figure*}

Understanding and leveraging spatial relations between objects is a desirable capability of service 
robots to function in human-centered environments. However, our environments are rich with everyday 
objects of various shapes and sizes, making it infeasible to pre-program a robot with sufficient 
knowledge to handle all arbitrary relations and objects it might encounter in the real world. 
Instead, we should equip robots with the ability to learn arbitrary relations in a lifelong manner 
and to generalize them to new objects, see Fig.~\ref{fig:SmallCover}. For example, having learned how to place a book inside a drawer, 
a robot should be able to generalize this spatial relation to place a toy inside a basket.\\
In this work, we propose a novel, neural-network-based approach to generalize spatial relations 
from the perspective of distance metric learning. Rather than considering a pre-specified set of 
relations and learning an individual model for each, our approach considers a continuous spectrum 
of pairwise relations and learns a metric that captures the similarities between scenes with 
respect to the relations they embody. Accordingly, we use this metric to generalize a relation to 
two new objects by minimizing the distance between the corresponding scenes in the learned metric 
as shown in Fig.~\ref{fig:Cover}. 
Following the metric-learning approach by Chopra~\etal~\cite{chopra2005siamese}, we use a variation 
of the siamese architecture~\cite{bromley1993siamesesignature} to train a convolutional 
neural network as a function that maps an input point cloud of a scene consisting of two objects to the 
metric space such that the Euclidean distance between points in that space captures the similarity 
between the spatial relations in the corresponding scenes. Our deep metric learning approach allows the 
robot to learn rich representations of spatial relations directly from point cloud input and without the 
need for manual feature design.
Furthermore, to generalize spatial relations in an end-to-end manner, we introduce a novel, gradient-descent based 
approach that leverages the learned distance metric to optimize the 3D poses of two objects in a scene in order to 
imitate an arbitrary relation between two other objects in a reference scene, see Fig.~\ref{fig:Cover}. For this, 
we backpropagate beyond the first convolution layer to optimize the translation and rotation of the object point clouds. 
Our gradient-based optimization enables the robot to imitate spatial relations based on visual demonstrations in an 
online and intuitive manner. 
In summary, we make the following contributions in this work: (1) an end-to-end approach to learning a metric for 
spatial relations from point clouds, (2) a differentiable projection to depth images to reduce the input dimensionality of point clouds, (3) a network architecture that models a differentiable metric function using 
a gradient approximation that allows for optimization beyond the first convolution layer, and (4) a demonstration 
that this technique enables gradient-based optimization in the learned feature space to optimize 3D translations 
and rotations of two new objects in order to generalize a demonstrated spatial relation.

\section{Related Work}
Learning spatial relations provides a robot with the necessary capability to carry out tasks 
that require understanding object interactions, such as object manipulation~\cite{zampogiannis2015actiondescriptors}, human-robot interaction~\cite{guadarrama2013grounding,shridhar2017grounding,
schulz2017collaborative} and active 
object search in complex environments~\cite{aydemir2011search}.   
In the context of robotics, learning spatial relations between objects has 
previously been phrased as a supervised classification problem based on 
handcrafted features such as contact points and relative poses~\cite{rosman2011learning,sjoo2011learning,fichtl2014spatialrelations3dhist}.
Spatial relations can also be learned from human-robot interaction using 
active learning, where the robot queries a teacher in order to refine a 
spatial model~\cite{kulick2013active}. However, the above techniques require 
learning an individual model for each relation and are thus limited in the 
number of relations they can handle. In contrast to these works, our metric 
learning approach allows us to reason about a continuous spectrum of known relations. 
Learning of object interactions from contact distributions has been addressed 
in the context of object grasping~\cite{kroemer2017kernel} and object 
placing~\cite{jiang2012learning}. While those approaches perform classification, 
our metric learning approach enables generation of spatial relations between 
objects solely based on visual information and without explicit modeling of 
contacts or physical object interaction. 
Visuospatial skill learning can imitate goal configurations
with objects based on spatial relations, but the imitation does not generalize to 
objects of various sizes and shapes~\cite{ahmadzadeh2017visuospatial}.

In our previous work we introduced a novel method that leverages large margin 
nearest neighbor metric learning in order to generalize spatial relations to 
new objects \cite{mees2017metric}. For this, we relied on hand-crafted 3D features 
to describe a scene. In contrast to this, we learn representations 
in an end-to-end fashion based on 2D image projections of scenes. Additionally, 
the previous work employed a grid search-based optimization with one object kept 
fixed whereas our approach optimizes on the full continuous spectrum of possible 
poses for both objects.

Our approach is related to deep learning techniques that learn similarity metrics 
directly from images, such as siamese~\cite{simo2015discriminative} and triplet networks~\cite{wang2014triplet}. 
In comparison, our network takes point clouds as input and 
processes them using our differentiable point cloud to depth image projection layer for 
input dimensionality reduction. In addition, we leverage the gradient of the 
metric to optimize the translation and rotation of objects in the point cloud space.
This strategy is in spirit similar to previous works that manipulate input images by 
backpropagating with respect to the input to visualize representations~\cite{simonyan2013deep}, 
and trick neural networks into making wrong classifications~\cite{szegedy2013intriguing}. 
We explore and analyze the utility of the metric's gradient for optimization of 3D translation 
and rotation in Section~\ref{results:generalizing}.

%
%
\section{Problem Formulation}
\label{sec:Problem}
We aim to learn a continuous representation for spatial relations between everyday objects in the 
form of a metric function, i.e., a representation that is not restricted to a finite set of relations, 
and to use this metric to enable a robot to imitate these spatial relations with new objects. Our goal is 
to learn this representation directly from geometric information in the form of raw point clouds, 
without requiring any semantics and without relying on handcrafted features. \\
For this, we consider pairwise spatial relations between objects. We denote these objects by $o_m$ and $o_n$ 
and we represent them as point clouds $\mathbf{P}_m$ and $\mathbf{P}_n$.
Together with the respective translation vectors $\mathbf{t}_m$, $\mathbf{t}_n$ expressed relative to a global world frame and rotation quaternions $\mathbf{q}_m$, $\mathbf{q}_n$, 
we define a scene $\mathbf{s}_i$ as the tuple
$\mathbf{s}_i=\langle o_{m,i}, o_{n,i}, \mathbf{t}_{m,i}, \mathbf{t}_{n,i}, \mathbf{q}_{m,i}, \mathbf{q}_{n,i}\rangle$. 
As a reference frame, we assume that the gravity vector $\mathbf{g}$ is known and oriented in the opposite 
direction of the global $z$-axis. 

To learn the metric, we require a set of training scenes 
$S = \{\mathbf{s}_0, ..., \mathbf{s}_n\}$ accompanied by labels in the form of a similarity matrix $\mathbf{Y}$ 
where the entry $\mathbf{Y}_{ij}$ denotes the similarity of the spatial relation between scenes $\mathbf{s}_i \in S$ 
and $\mathbf{s}_j \in S$. That is $\mathbf{Y}_{ij}$ should be small for similar relations and large for dissimilar 
relations. Note that we do not require all possible scene combinations to be labeled, i.e., $\mathbf{Y}$ does not 
need to be fully specified. To ease labeling, we allow the entries of $\mathbf{Y}$ to be binary, i.e., $\mathbf{Y}_{ij}\in\{0,1\}$, where $0$ means similar and $1$ dissimilar.
\\
%
%
%
Our goal is to learn a metric function $f(\mathbf{s}_i, \mathbf{s}_j) = d$ that maps two scenes to
a distance $d$ such that the following properties hold: (1) $d$ captures the
similarity of the spatial relations depicted in scenes $\mathbf{s}_i$ and $\mathbf{s}_j$, that is
$d$ is small for similar relations and large for dissimilar relations, and (2) $f$
is differentiable. The latter ensures that we can employ gradient based optimization on the metric function. \\
Instead of directly learning the metric function $f$, we learn a mapping function $\Gamma$ that maps each 
input scene into a low-dimensional space such that the Euclidean distance captures the similarity
of spatial relations. 
Concretely, we define our metric function $f$ as
\begin{align} \label{eq:metric}
f(\mathbf{s}_i, \mathbf{s}_j) = ||\Gamma(\mathbf{s}_i) - \Gamma(\mathbf{s}_j)||_2.
\end{align}

As a smaller distance denotes higher similarity, we formulate the problem of generalizing a spatial relation 
in a reference scene $\mathbf{s}_r$ to a test scene $\mathbf{s}_t$ as finding the translations and rotations
of both objects in $\mathbf{s}_t$ which minimize the distance between the two scenes under the learned metric, i.e., we seek to solve the following problem:
\begin{align} \label{eq:optproblem}
\minimize_{\substack{\mathbf{t}_{m,t}, \mathbf{t}_{n,t}, \mathbf{q}_{m,t}, \mathbf{q}_{n,t}}}{f(\mathbf{s}_r, \mathbf{s}_t)}.
\end{align} 
In this work, we focus on computing the poses of both objects to imitate the semantics of a reference relation, and do not consider the physical feasibility of the resulting scene, e.g., collision checks.
%
%
%
\section{Approach}
\label{sec:approach}
Our method consists of two phases.
In the first, we learn a distance metric function to capture the semantics of spatial relations. In the second, we leverage the gradient of this function to imitate spatial relations in a continuous manner via optimization of the object poses. The key challenges are to learn a rich representation from high dimensional input and to backpropagate the gradient information into the raw point cloud.\\
\subsection{Distance Metric Learning: Metric Composition and Training}
\begin{figure}[t]
\begin{centering}
	\tdplotsetmaincoords{60}{125}
    \scalebox{0.8}{
	\begin{tikzpicture}[tdplot_main_coords,
				        cube/.style={very thick,black},
				        axis/.style={->,aisblue,thick}]

		\draw[axis] (-0.5,-0.5,0) -- (2.5,-0.5,0) node[anchor=west]{$x$};
		\draw[axis] (-0.5,-0.5,0) -- (-0.5,2.5,0) node[anchor=west]{$y$};
		\draw[axis] (-0.5,-0.5,0) -- (-0.5,-0.5,4) node[anchor=west]{$z$};

		\draw[axis] (-0.2,-2,2.2) -- (1.8,-2,2.2) node[anchor=south, xshift=0.1cm, yshift=0.1cm]{$x_\mat{U}$};
		\draw[axis] (-0.2,-2,2.2) -- (-0.2,-2,0.6) node[anchor=west]{$y_\mat{U}$};

	    \node at (-0.4,-3,0) {\includegraphics[scale=1.3]{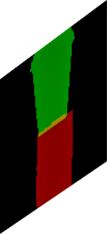}};
	    \node at (-3.5,0,0) {\includegraphics[scale=1.3]{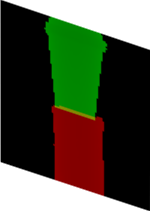}};

	    \node at (2.0, -2.4, 1) {$\displaystyle ^1\mat{U}$};
        \node at (-2, 2.4, 1) {$\displaystyle ^2\mat{U}$};
        \node at (1.8, 0.3, 4.3) {$^0\mat{U}$};

		\draw[densely dotted,very thick,gray] (0,0,2) -- (0,0,3.5);
		\draw[densely dotted,very thick,gray] (0,2,2) -- (0,2,3.5);
		\draw[densely dotted,very thick,gray] (2,0,2) -- (2,0,3.5);
		\draw[densely dotted,very thick,gray] (2,2,2) -- (2,2,3.5);

		\draw[densely dotted,very thick,gray] (0,0,0) -- (0,-2,0);
		\draw[densely dotted,very thick,gray] (2,0,0) -- (2,-2,0);
		\draw[densely dotted,very thick,gray] (0,0,2) -- (0,-2,2);
		\draw[densely dotted,very thick,gray] (2,0,2) -- (2,-2,2);
		
	    \draw[densely dotted,very thick,gray] (0,0,0) -- (-2,0,0);
		\draw[densely dotted,very thick,gray] (0,2,0) -- (-2,2,0);
		\draw[densely dotted,very thick,gray] (0,0,2) -- (-2,0,2);
		\draw[densely dotted,very thick,gray] (0,2,2) -- (-2,2,2);
	    \node at (2.4,2,4.5) {\includegraphics[scale=1.3]{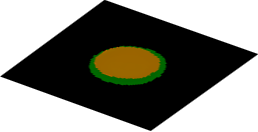}};

		\draw[cube] (0,0,0) -- (0,2,0) -- (2,2,0) -- (2,0,0) -- cycle; 
		
		\draw[cube] (0,0,0) -- (0,0,2);
		\draw[cube] (0,2,0) -- (0,2,2);
		\draw[cube] (2,0,0) -- (2,0,2);

	    \node at (-0.3,0,0) {\includegraphics[scale=0.25]{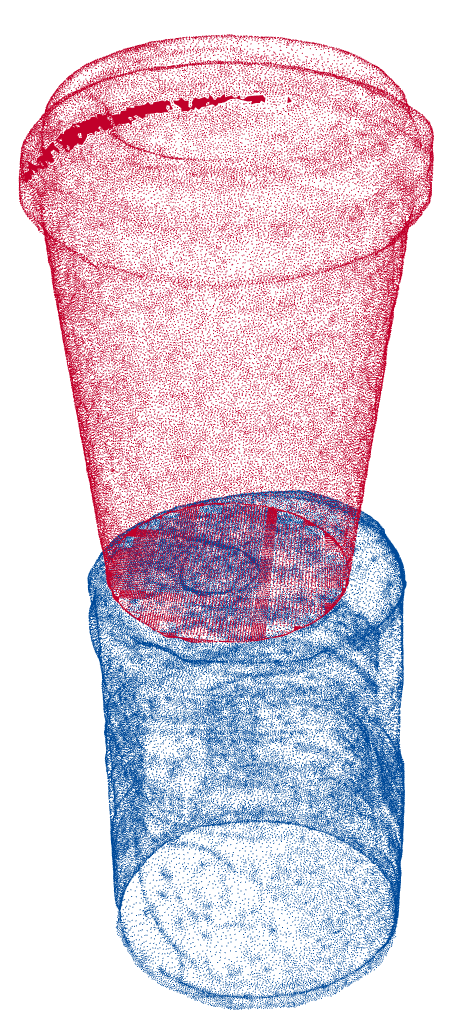}};
	    \draw[cube] (0,0,2) -- (0,2,2) -- (2,2,2) -- (2,0,2) -- cycle;  
	    \draw[cube] (2,2,0) -- (2,2,2);  

	    \node (xy) [scale=2] at (0, -2.8, -4) {\includegraphics{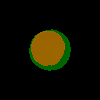}};
	    \node (xz) [scale=2,right=of xy, xshift=-0.6cm] {\includegraphics{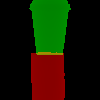}};
	    \node (zy) [scale=2,right=of xz, xshift=-0.6cm] {\includegraphics{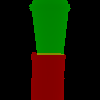}};
        \node [below=of xy, yshift=1cm] {$z-1 = 0$};
        \node [below=of xz, yshift=1cm] {$y = 0$};
        \node [below=of zy, yshift=1cm] {$x = 0$};

    \end{tikzpicture}
    }

    \caption[Projections to three orthogonal planes]{Projections to three orthogonal planes. We project each point cloud to the three orthogonal planes defined by $y = 0$, $x = 0$, and $z-1 = 0$. We create a depth image by setting the value of a pixel to the smallest distance of all points that are projected on this pixel multiplied by 100. To contrast the objects from the background we add a bias of 100 to all object pixels.
    Each projection of the two objects is in a separate channel and we randomly choose which object is positioned in the first channel. For this visualization, we added an all-zero third channel.}
    \label{fig:Projections}
\end{centering}
 \end{figure}
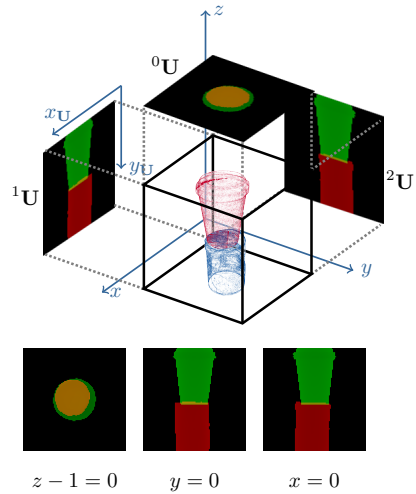

The goal of learning the distance metric is to express the similarity between spatial relations. As input 
we use high-dimensional 3D point cloud data.\\
Therefore we seek a mapping function $\Gamma$ that reduces the dimensionality from the point cloud space 
to a low-dimensional metric space. We implement $\Gamma$ as a composition of three functions, which we 
will now outline briefly and then describe two of the functions more detailed. First, the transformation
function $\psi$ applies the corresponding rotation $\mathbf{q}$ and translation $\mathbf{t}$ to each
object point cloud.
Second, we project the point clouds to three orthogonal planes to create three depth images. We denote 
this projection function by $\rho$. Third, we apply a mapping function $G_\mathbf{W}$ parameterized 
by $\mathbf{W}$, which maps the three projections to the metric space, see Fig.~\ref{fig:Cover} for 
an overview. More formally, we compose the mapping $\Gamma$ as
\begin{align}\label{eg:composition}
\Gamma \coloneqq G_{\mathbf{W}} \circ \rho \circ \psi.
\end{align}\\
%
%
We tackle the dimensionality of the input data using a function $\rho$ that projects 
the point clouds to 
three depth images. This serves as a non-parameterized reduction of the input dimensionality in comparison 
to 3D representations such as octrees or voxels. 
Concretely, we scale the scene to fit in a unit cube, see Fig.~\ref{fig:Projections}. We then project 
each point to three orthogonal image planes of size $100\times100$ pixels fit to the top, front, and 
side faces of the cube such that the image plane normals are either parallel or orthogonal to the gravity
vector $\mathbf{g}$. We place the projection of each object in a separate channel. In this work, we will 
refer to a single orthogonal projection of one object as projection image.
%
%
\\
To learn the parameters $\mathbf{W}$ of the mapping function $G_\mathbf{W}$ we use a triplet network~\cite{wang2014triplet}, a variation of the
siamese convolutional network that features three identical, weight-sharing networks $G_\mathbf{W}$.
We train the network on input triplets of projections
$\langle (\rho \circ \psi)(\mathbf{s}_i), (\rho \circ \psi)(\mathbf{s}^+_j), (\rho \circ \psi)(\mathbf{s}^-_k) \rangle$ where $\mathbf{s}_i \in S$ is
a reference scene and  $\mathbf{s}^+_j, \mathbf{s}^-_k\in S$ are similar and dissimilar to $\mathbf{s}_i$, respectively, that is, in the case of binary labels, $\mathbf{Y}_{ij} = 0$ and $\mathbf{Y}_{ik} = 1$.
We run each plane of the projection through its own sub-network with the sub-networks also sharing the weights and 
we fuse them with a fully-connected layer, see Fig.~\ref{fig:Architecture} for architecture 
details.

In contrast to an actual triplet network we do not employ a ranking loss but adapt the hinge
loss function as in the approach by Chopra~\etal~to enforce an upper bound on the distance \cite{chopra2005siamese}. This upper bound ensures that the learning rate used for optimizing the poses 
of a test scene can be tuned independently of a specific set of learned parameters $W$.
Concretely, we compute the loss function
\begin{align}
C\big(\Gamma(\mathbf{s}), \Gamma(\mathbf{s}^+), \Gamma(\mathbf{s}^-)\big) = \frac{1}{2}(d_+)^2 + \frac{1}{2}\big(\max(0, 1 - d_-)\big)^2,
\end{align}
where $\Gamma(\mathbf{s})$ denotes the embedding of the scene $\mathbf{s}$, i.e., ${\Gamma(\mathbf{s}) = G_\mathbf{W}\big((\rho \circ \psi)(\mathbf{s})\big)}$,
and $d_+$, $d_-$ denote the Euclidean distance of ${\Gamma(\mathbf{s}^+)}$ and ${\Gamma(\mathbf{s}^-)}$ to the embedding ${\Gamma(\mathbf{s})}$ of the reference scene, respectively. During optimization, this results in $d_+$ being minimized towards $0$ and $d_-$ being maximized towards $1$.
\newcommand{\conv}[6] {
  \node (rect) at (#1,#2) [draw=black!#5, color=black!#5, minimum width=4cm,minimum height=0.5cm, inner sep=1pt]
        {$#3 \times#3$ conv, $#4$, elu#6};
}

\newcommand{\sibling}[5] {
    \node (s1) at (#1,#2) [draw=aisblue, thick, color=aisblue, minimum width=0.35cm,minimum height=0.8cm, inner sep=1pt] {};
    \node (s2) at (#1+0.45,#2) [draw=aisblue, thick, color=aisblue, minimum width=0.35cm,minimum height=0.8cm, inner sep=1pt] {};
    \node (s3) at (#1+0.9,#2) [draw=aisblue, thick, color=aisblue, minimum width=0.35cm,minimum height=0.8cm, inner sep=1pt] {};
    \node (fc1) at (#1+0.45,#2-1.0) [draw=black, thick, color=black, minimum width=1.2cm,minimum height=0.5cm, inner sep=1pt] {fc, 64};
    \draw[->, >=stealth, draw=black] (s1.south) -- (fc1.north -| s1.south);
    \draw[->, >=stealth, draw=black] (s2.south) -- (fc1.north -| s2.south);
    \draw[->, >=stealth, draw=black] (s3.south) -- (fc1.north -| s3.south);
    \node (p1) [above of=s1, yshift=0cm, inner sep=0pt] {\includegraphics[width=0.4cm, interpolate=false]{#3}};
    \node (p2) [above of=s2, yshift=0cm, inner sep=0pt] {\includegraphics[width=0.4cm, interpolate=false]{#4}};
    \node (p3) [above of=s3, yshift=0cm, inner sep=0pt] {\includegraphics[width=0.4cm, interpolate=false]{#5}};
    \draw[->, >=stealth] (p1.south) -- (s1.north);
    \draw[->, >=stealth] (p2.south) -- (s2.north);
    \draw[->, >=stealth] (p3.south) -- (s3.north);

}

\begin{figure}[t!]
\begin{center}
    \scalebox{0.8}{
    \begin{tikzpicture}[framed,background rectangle/.style={thick, draw=aisblue}]
    \node (rect) at (0,0.8) [draw,minimum width=3.4cm,minimum height=0.5cm, inner sep=1pt, align=center] {input: $2 \times 100 \times 100$};
    \draw[->, >=stealth] (rect.south) -- (0, 0.25);
    \conv{0}{0}{10}{32}{100}{, pool}
    \draw[->, >=stealth, draw=black] (rect.south) -- (0, -0.45);
    \conv{0}{-0.7}{8}{42}{100}{, pool}
    \draw[->, >=stealth, draw=black] (rect.south) -- (0, -0.7-0.45);
    \conv{0}{-1.4}{6}{64}{100}{, pool}
    \draw[->, >=stealth, draw=black] (rect.south) -- (0, -1.4-0.45);
    \conv{0}{-2.1}{4}{64}{100}{, pool}
    \draw[->, >=stealth, draw=black] (rect.south) -- (0, -2.1-0.45);
    \conv{0}{-2.8}{4}{128}{100}{, pool}
    \draw[->, >=stealth, draw=black] (rect.south) -- (0, -2.8-0.45);
    \conv{0}{-3.5}{4}{128}{100}{}
    \draw[->, >=stealth, draw=black] (rect.south) -- (0, -3.5-0.45);
    \conv{0}{-4.2}{2}{128}{100}{}

    \end{tikzpicture}
    }
    \begin{tikzpicture}
 
    \node (l2) at (1.3, -4.1) [circle, inner sep=0pt, draw=black] {$d_+$};
    \node (l22) at (2.9, -4.1) [circle, inner sep=0pt, draw=black] {$d_-$};
    \sibling{0}{-2.4}{figures/projection/beans_coffeMensa_9570_xy.png}{figures/projection/beans_coffeMensa_9570_xz.png}{figures/projection/beans_coffeMensa_9570_yz.png};
    \draw (fc1.255) -- (l2.west -| fc1.255) -- (l2.west);
    \sibling{1.6}{-2.4}{figures/cover/beansbowlSal7441xy.png}{figures/cover/beansbowlSal7441xz.png}{figures/cover/beansbowlSal7441yz.png};;
    \draw (fc1.255) -- (l2.east -| fc1.255) -- (l2.east);
    \draw (fc1.285) -- (l22.west -| fc1.285) -- (l22.west);
    \sibling{3.2}{-2.4}{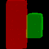}{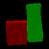}{figures/cover/proj0_1.png};;
    \draw (fc1.285) -- (l22.east -| fc1.285) -- (l22.east);
    \node (loss) at (2.1, -4.9) [draw=black] {training loss};
    \draw[->, >=stealth] (l2.south) -- (loss.north -| l2.south);
    \draw[->, >=stealth] (l22.south) -- (loss.north -| l22.south);

    \end{tikzpicture}
\end{center}

 \caption{The hyperparameters of the subnet of a sibling $G_\mathbf{W}$, found
    with random search. Each subnet receives one projection of one plane. 
    The convolution layers of each subnetwork share the
    weights. All three subnets are fused with a fully connected layer.
    The sibling network $G_\mathbf{W}$ is then cloned three times into a triplet
    network when training the distance metric and cloned two times into a
    siamese network when generalizing a relation at test time.}
\label{fig:Architecture}
\end{figure}
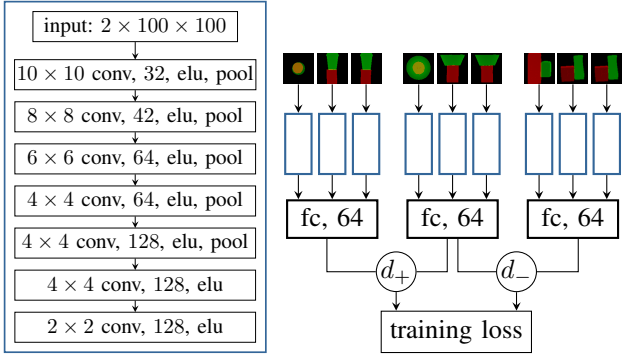
\label{sec:approachgradient}
\subsection{Generalizing Spatial Relations Using the Backward Pass}
\begin{figure}[t]
\begin{centering}
	\tdplotsetmaincoords{60}{125}
    \scalebox{0.7}{
	\begin{tikzpicture}[tdplot_main_coords,
				        cube/.style={very thick,black},
				        axis/.style={->,aisblue,thick},
				        >=stealth]

		\draw[axis] (-0.5,-0.5,0) -- (2.5,-0.5,0) node[anchor=west]{$x$};
		\draw[axis] (-0.5,-0.5,0) -- (-0.5,2.5,0) node[anchor=west]{$y$};
		\draw[axis] (-0.5,-0.5,0) -- (-0.5,-0.5,4) node[anchor=west]{$z$};

		\draw[axis] (-0.2,-2,2.2) -- (1.8,-2,2.2) node[anchor=south, xshift=0.1cm, yshift=0.1cm]{$x_\mat{U}$};
		\draw[axis] (-0.2,-2,2.2) -- (-0.2,-2,0.6) node[anchor=west]{$y_\mat{U}$};

	    \node at (-0.4,-4.5,0) {\includegraphics[scale=1.3]{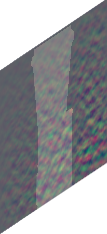}};
	    \node at (2.3, -4.7, 1) {$\displaystyle ^1\mat{U}_y' =  \frac{\partial (^1\mat{U}')}{\partial y_\mat{U}}$};
	    \node at (-0.4,-3,0) {\includegraphics[scale=1.3]{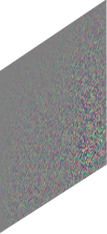}};
	    \node at (2.0, -2.4, 0.5) {$\displaystyle ^1\mat{U}'$};
	    \node at (-5,0,0) {\includegraphics[scale=1.3]{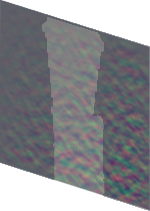}};
	    \node at (-4.5, 2.9, 1) {$\displaystyle ^2\mat{U}'_y =  \frac{\partial (^2\mat{U}')}{\partial y_\mat{U}}$};
	    \node at (-3.5,0,0) {\includegraphics[scale=1.3]{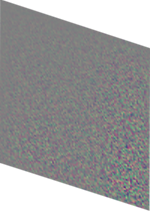}};
	    \node at (-1.6, 1.4, 0) {$\displaystyle ^2\mat{U}'$};
		\draw[densely dotted,very thick,gray] (0,0,2) -- (0,0,3.5);
		\draw[densely dotted,very thick,gray] (0,2,2) -- (0,2,3.5);
		\draw[densely dotted,very thick,gray] (2,0,2) -- (2,0,3.5);
		\draw[densely dotted,very thick,gray] (2,2,2) -- (2,2,3.5);

		\draw[densely dotted,very thick,gray] (0,0,0) -- (0,-2,0);
		\draw[densely dotted,very thick,gray] (2,0,0) -- (2,-2,0);
		\draw[densely dotted,very thick,gray] (0,0,2) -- (0,-2,2);
		\draw[densely dotted,very thick,gray] (2,0,2) -- (2,-2,2);
		
	    \draw[densely dotted,very thick,gray] (0,0,0) -- (-2,0,0);
		\draw[densely dotted,very thick,gray] (0,2,0) -- (-2,2,0);
		\draw[densely dotted,very thick,gray] (0,0,2) -- (-2,0,2);
		\draw[densely dotted,very thick,gray] (0,2,2) -- (-2,2,2);
	    
	    \node at (2.4,2,4.5) {\includegraphics[scale=1.3]{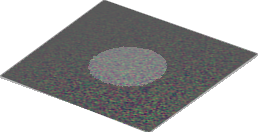}};
	    \node at (2.4,2,6.0) {\includegraphics[scale=1.3]{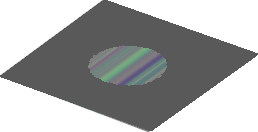}};
	    \node at (2.4,2,7.5) {\includegraphics[scale=1.3]{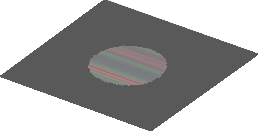}};
	    \node at (2.3, -0.1, 3.9) {\Huge{+}};
	    \node at (2.3, -0.1, 5.4) {\Huge{+}};
	    \node at (2.0, 0.3, 4.2) {$^0\mat{U}'$};

	    \draw[->, very thick] (2.1, -2, 0) to node[sloped, anchor=center, below] {$\mat{S}_y \ast$} (2.1, -3.5, 0);
	    \node (xz_arrow) at (-0.4,-4.5,0.9) {};
	    \node (xz_arrow_end) at (3.3,1,7.5) {};
	    \draw[->, very thick] (xz_arrow) edge[bend left] node [left] {$\sum_{y_\mat{U}}(^1\mat{U}'_y)$} (xz_arrow_end);

	    \node (yz_arrow) at (-3.5,1,2) {};
	    \node (yz_arrow_end) at (-2,0.7,3.7) {};
	    \draw[->, very thick] (yz_arrow) edge[bend right] node [right] {$\sum_{y_\mat{U}}(^2\mat{U}'_y)$} (yz_arrow_end);

	    \draw[->, very thick] (-2, 2.1, 0) to node[sloped, anchor=center, below] {$\mat{S}_y \ast$} (-3.5, 2.1, 0);

		\draw[cube] (0,0,0) -- (0,2,0) -- (2,2,0) -- (2,0,0) -- cycle; 
		
		\draw[cube] (0,0,0) -- (0,0,2);
		\draw[cube] (0,2,0) -- (0,2,2);
		\draw[cube] (2,0,0) -- (2,0,2);

	    \node at (-0.3,0,0) {\includegraphics[scale=0.25]{figures/projection/coffeebeanspointcloud.png}};
	    \draw[cube] (0,0,2) -- (0,2,2) -- (2,2,2) -- (2,0,2) -- cycle;  
	    \draw[cube] (2,2,0) -- (2,2,2);  

    \end{tikzpicture}
    }

    \caption[Partial Derivative Implementation]{Implementation of the partial derivative w.r.t. the world $z$-axis. For each projection we compute the partial derivatives w.r.t. pixels, $\mat{U}'$. For both projections $^1\mat{U}, ^2\mat{U}$ the world $z$-axis corresponds to the image axis $y_\mat{U}$. We compute the partial derivatives w.r.t. $y_\mat{U}$ by convolving  $^1\mat{U}', ^2\mat{U}'$ and the Sobel kernel $\mat{S}_y$. We then sum the resulting errors over the $y_\mat{U}$-axis and add them to the top projection, propagated over the axis of $^0\mat{U}$ that corresponds to the resepctive depth axis of $^1\mat{U}, ^2\mat{U}$. Then, for each pixel, we assign this error to the $z$-coordinate of the closest point the pixel in $^0\mat{U}$ originated from. Note that, assuming a solid point cloud with uniformly-distributed points, this is equivalent to the computation described in \ref{sec:approachgradient}.

    }
    \label{fig:GradientDiagram}
\end{centering}
 \end{figure}
\label{sec:approachBackwarPass}
Having learned the neural network mapping function $G_\mat{W}$, we can now leverage the backpropagation algorithm to imitate a relation by optimizing the parameters of the 3D transformations applied to the point clouds. As stated in~(\ref{eq:optproblem}), we formulate the generalization of a spatial relation as a minimization problem with respect to the rotations and translations in the scene. Note that this differs from the representation learning process in that we keep the parameters $\mat{W}$ fixed and instead optimize the transformation parameters $\mathbf{t},\mathbf{q}$.\\
To employ gradient based optimization, the transformation $\Gamma$ must be differentiable. While the functions $G_{\mathbf{W}}$ and $\psi$ are differentiable, the gradient of the projection $\rho$ needs a more thorough consideration. 
When backpropagating through the input layer of the first convolutional operation of $G_\mat{W}$, we need to consider that
an input pixel not only contains depth information, but also discrete spatial information.
Projecting a point onto the image plane discretizes two dimensions,
which makes gradient-based optimization on these two axes impractical.
Although projecting a scene to three sides sustains one continuous gradient for each axis, in our
application the important information is contained in the location of the pixel, i.e., in the discretized
dimensions.

As an example, consider a top-view projection image $^0\mat{U}$ of a `cup on top of a can', i.e., a pixel value $u_{y,x}$ corresponds to the $z$-value of the point in the world frame, see Fig.~\ref{fig:GradientDiagram}. With only the depth information one cannot conclude if the cup is resting on the can or if it is hovering above it, as no information of the bottom of the cup is available. In contrast, the side view captures this information on the $y_\mat{U}$-axis of the image which also corresponds to the $z$-axis in the world coordinate frame. However, the gradient of the loss with respect to $y_\mat{U}$ is not well defined. 
The crux here is that the function $G_\mat{W}$, expressed as a convolutional neural network, only computes partial derivatives with respect to the input $\frac{\delta C}{\delta u_{y,x}}$, i.e., the partial derivative only depends on the magnitude $u_{y,x}$ but not on the position of a pixel. However, the hierarchical structure of $G_\mat{W}$ retains the spatial context of the error, which we will use to propagate the error of a single projection image back to all three coordinates of the 3D points.

Therefore, we convolve the matrix containing the error with respect to the input image of $G_\mat{W}$ with a Sobel-derived kernel. This procedure approximates the rate of change of the error of the input image with respect to $x_\mat{U}$ and with respect to $y_\mat{U}$. That is, we approximate the change of the error with respect to shifting a pixel on the $x_\mat{U}$ and $y_\mat{U}$ axis and use this as the error of the respective spatial coordinates of the pixel. This error can then be backpropagated to the 3D point it resulted from via applying the inverse orthogonal projection and all three coordinates of the point are updated. \\
More formally, for a projection image $\mat{U}$, i.e., one input channel of the function $G_\mat{W}$, the partial derivative can be formulated as follows. Let $\mat{U}'$ be the gradient of this projection image with respect to the loss where the entry $u'_{y,x} = \frac{\delta C}{\delta u_{y,x}}$ denotes the partial derivative of the loss with respect to the input pixel at position $(x, y)$. We compute the matrices of partial derivatives with respect to the $y$ and $x$ position $\mat{U}'_y$ and $\mat{U}'_x$ as $\mat{U}'_y = \mat{S}_y \ast \mat{U}'$ and $\mat{U}'_x = \mat{S}_x \ast \mat{U}'$ with $\mat{S}_y, \mat{S}_x$ being the Sobel kernels

${\mat{S}_y= \left[ \begin{smallmatrix}
   1 & 2 & 1 \\
   0 & 0 & 0 \\
   -1 & -2 & -1 \\
   \end{smallmatrix}
   \right]}$
and
${\mat{S}_x= \left[ \begin{smallmatrix}
   1 & 0 & -1 \\
   2 & 0 & -2 \\
   1 & 0 & -1 \\
   \end{smallmatrix}
   \right]}$.
In practice, we found that using a larger Sobel-derived kernel to approximate the derivatives achieves better results, which likely results from the gradients being sparse due to maxpooling.
Once we have computed an error on the pixel $u_{y,x}$ with respect to $x_\mat{U}$ and $y_\mat{U}$ we assign the error values to the respective coordinates of all points whose projection would result in the pixel $y,x$, assuming the object is solid. For each point, we sum all the errors associated with it. In summary, this ensures that each coordinate of a point is assigned an error from each projection. The remaining partial derivatives for the translation and rotation have an analytical solution.
Fig.~\ref{fig:GradientDiagram} depicts an overview of the gradient implementation for a single axis of the points. Our code is available at \href{https://github.com/philjd/generalize_spatial_relations}{\url{https://github.com/philjd/generalize\_spatial\_relations}.
}
%
%
\section{Experimental Results}
\label{sec:result}

\newcommand{\gen}[4] {
    \node (ref) at (0,#1) [minimum width=1cm,minimum height=1cm] {
            \includegraphics[height=1cm, keepaspectratio, interpolate=false]{figures/experiments_overview/#2}};

    \node (init) [minimum width=1cm,minimum height=1cm, right=of ref] {
            \includegraphics[height=1cm, keepaspectratio, interpolate=false]{figures/experiments_overview/#3}};

    \node (gen) [minimum width=1cm,minimum height=1cm, right of =init, xshift=1.8cm] {
            \includegraphics[height=1cm, keepaspectratio, interpolate=false]{figures/generalization/#4}};
    \draw[->, >=latex] (init) -- (gen);

}

\begin{figure}[!t]
\begin{center}
    \subfloat[successful generalizations\label{fig:genexamplesuccess}]{
    \scalebox{0.85}{
    \begin{tikzpicture}[every node/.style = {anchor= north west}]
        \gen{0}{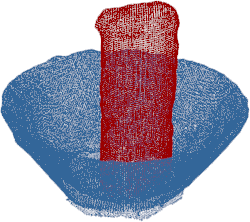}{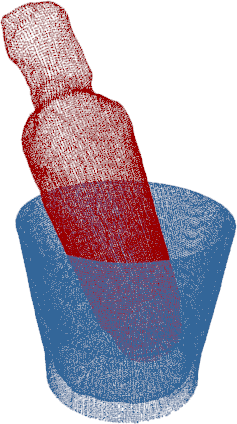}{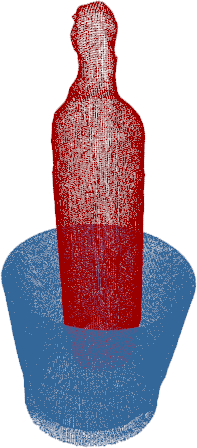}
        \node (reftext) [above=10pt of ref] {Reference};
        \node (inittext) [above=10pt of init] {Initial};
        \node (gentext) [above=10pt of gen] {Generalized};
        \draw (-0.3, 0.1) -- (6.3, 0.1);
        \gen{-1.7}{6_pot4_oil_3767.png}{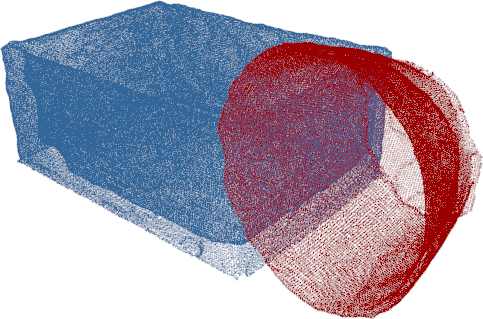}{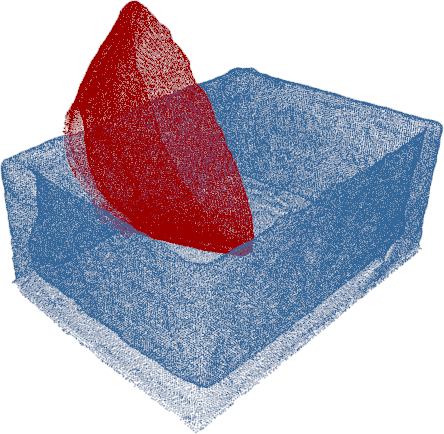}
        \gen{-3.5}{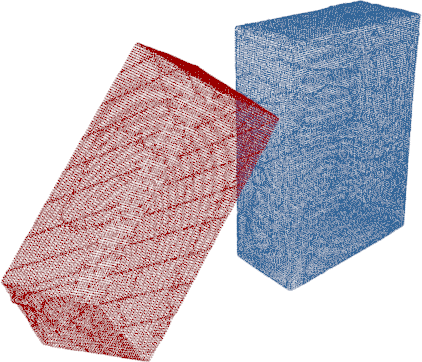}{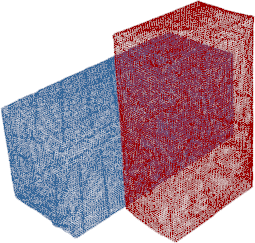}{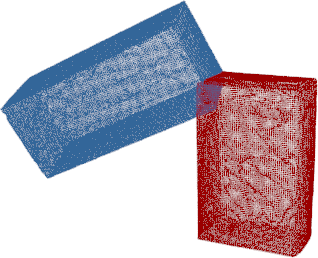}
    \end{tikzpicture}
    }
    } \\
    
    \subfloat[successful, but physically less reasonable generalizations\label{fig:genexampleunstable}]{
    \scalebox{0.85}{
    \begin{tikzpicture}
        \gen{0}{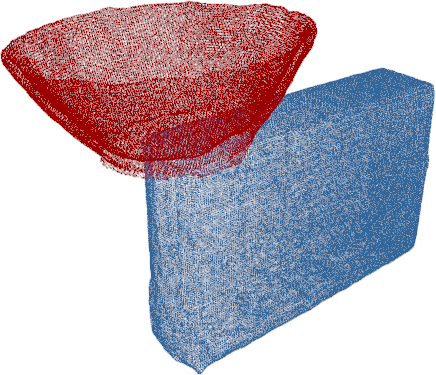}{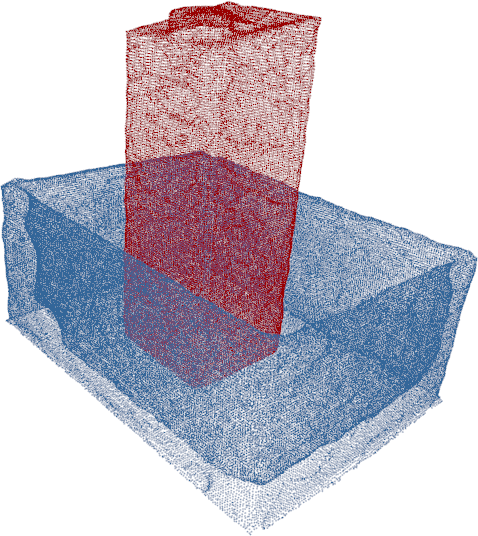}{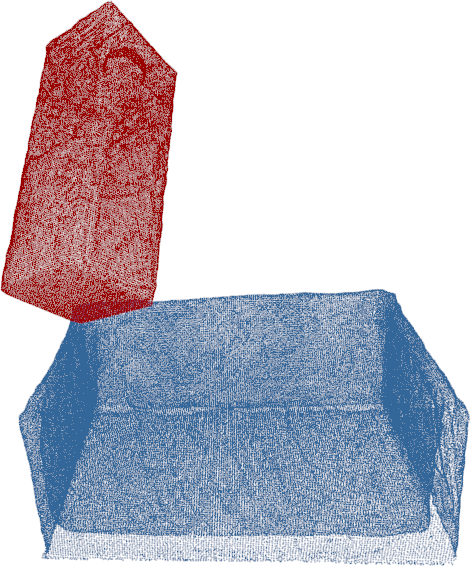}
        \gen{-1.7}{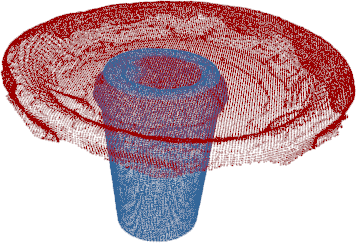}{3_rice_tea1_3271.png}{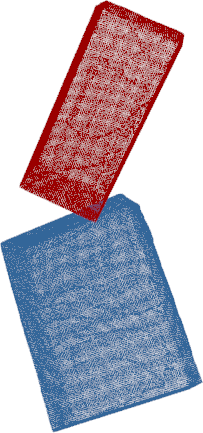}
        \gen{-3.5}{0_coffeMensa_colorPlate_8335.png}{3_SmallBoxBase_bowl_4380.png}{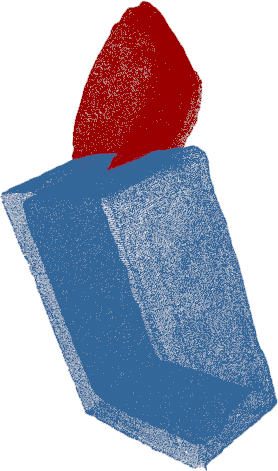}
    \end{tikzpicture}
    }
    } \\
    
    \subfloat[unsuccessful generalizations\label{fig:genexamplefailed}]{
    \scalebox{0.85}{
    \begin{tikzpicture}
        \gen{0}{1_bowl_sweetener_3349.png}{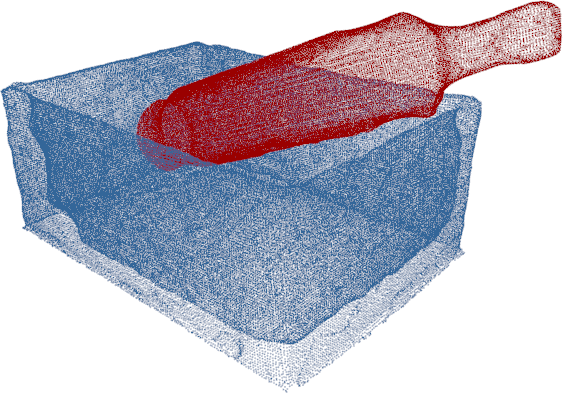}{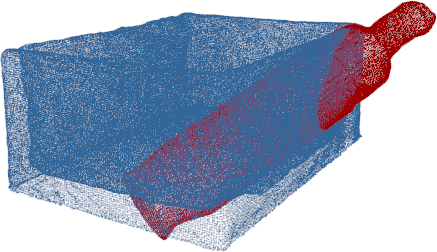}
        \gen{-1.7}{3_rice_tea1_3271.png}{6_pot4_oil_3767.png}{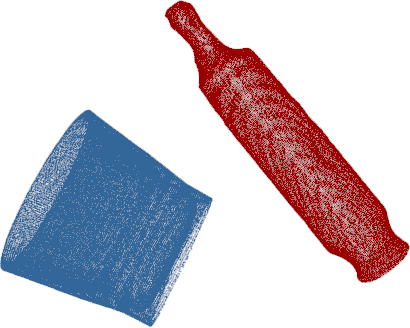}
        \gen{-3.5}{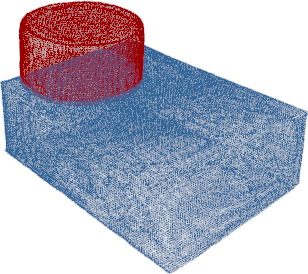}{3_rice_tea1_3271.png}{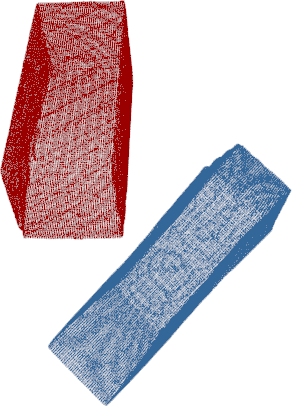}
    \end{tikzpicture}
    }
    }

\end{center}

 \caption{Examples for successful (Fig.~\ref{fig:genexamplesuccess}), successful but physically infeasible (Fig.~\ref{fig:genexampleunstable}),
          and unsuccessful generalizations (Fig.~\ref{fig:genexamplefailed}). In each row, the leftmost scene depicts the reference scene,
          the test scene before optimizing, and the rightmost scene depicts the generalized result.}
\label{fig:Generalization}
\end{figure}
\begin{table}[t]
  \centering
  \begin{tabular}{c c c}
  Method & 3-out-of-5 acc. &  5-out-of-5 acc. \\
  \hline
  \hline
  LMNN~\cite{mees2017metric} & $86.52\% \pm 1.98$ &  N/A \\
  GBLMNN~\cite{mees2017metric} & $87.6\% \pm 1.94$ & N/A \\
  Our NN-based metric & {$\mathbf{91.21\% \pm 2.78\%}$ } & $76.25\% \pm 7.21\%$\\
  \end{tabular}
  \caption{Nearest neighbor performance on the Freiburg Spatial Relations dataset. We report results for correctly retrieving 3-out-of-5 and 5-out-of-5 target neighbors of a query scene.}
  \label{tab:nnresults}
\end{table}

In this section we conduct several quantitative and qualitative experiments to benchmark the performance of our approach.
Hereby, we demonstrate the following: 1) our learned feature representation is able to generalize over a rich set of different spatial relations and yields improved performance for a spatial nearest neighbor retrieval task with respect to a state-of-the-art method,
2) using our novel gradient-based optimization method we are able to generalize spatial relations to new objects and to capture the intention of the reference scenes being imitated without prior semantic knowledge about the relation they embody.

\begin{figure*}[!tp]
    \begin{center}
    \begin{tikzpicture}[very thin]
        \node (refimg) at(-2.2,0) {\includegraphics[interpolate=false,height=1cm]{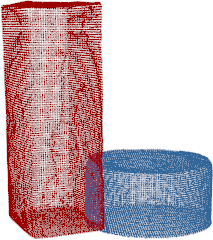}};
        \node (reftext) [below=0pt of refimg, align=center] {Reference};

        \node (optimg0) at(0.6, -0.4) [draw=black!40, thin, minimum width=1cm] {\includegraphics[height=1cm,keepaspectratio,interpolate=false]{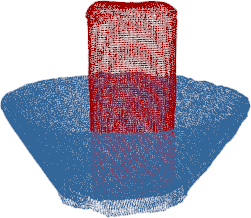}};
        \node (optimg1) at(3.7, -0.4) [draw=black!40, thin, minimum width=1cm] {\includegraphics[height=1cm,keepaspectratio,interpolate=false]{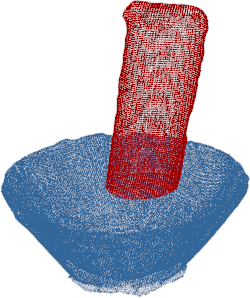}};
        \node (optimg2) at(6.2, -0.4) [draw=black!40, thin, minimum width=1cm] {\includegraphics[height=1cm,keepaspectratio,interpolate=false]{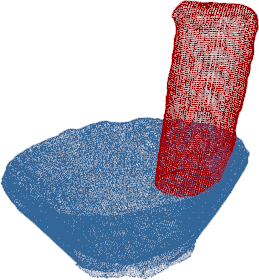}};
        \node (optimg4) at(8.7, -0.4) [draw=black!40, thin, minimum width=1cm] {\includegraphics[height=1cm,keepaspectratio,interpolate=false]{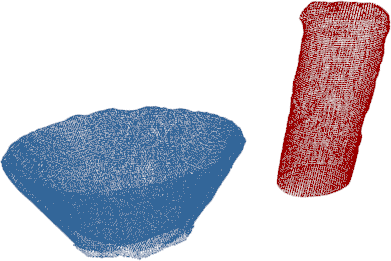}};
        \node (optimg5) at(12, -0.4) [draw=black!40, thin, minimum width=1cm] {\includegraphics[height=1cm,keepaspectratio,interpolate=false]{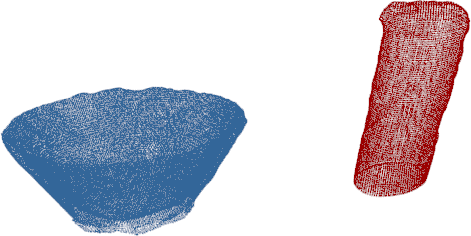}};
        
        \node (graph) at (6.5,-0.75) {
            \begin{tikzpicture}
            \begin{axis}[width=\textwidth-2cm, height=3cm, ytick={0,0.1,0.2}, enlargelimits=false,
                         every axis x label/.style={
                            at={(current axis.south)},below=2mm},
                         xlabel={steps}]
            \addplot+[mark=none, thick] table [x=step, y=distance, col sep=comma] {figures/optimization/ex2/dists.txt};
            \end{axis}
            \end{tikzpicture}
        };
        \node (d) [left=0pt of graph, yshift=1cm, xshift=0cm, rotate=90] {distance};
    \end{tikzpicture}
    \end{center}
    \caption{This figure shows an example of an optimization run to imitate a `next-to' relation in a reference scene consisting of a box and a can and to generalize it using two new objects: a bowl and a smaller box. We show the intermediate scenes obtained during the optimization and their corresponding distance to the reference scene. Despite the very different shapes, the produced relation resembles the spatial relation of the reference scene.}
    \label{fig:Optimization}
\end{figure*}
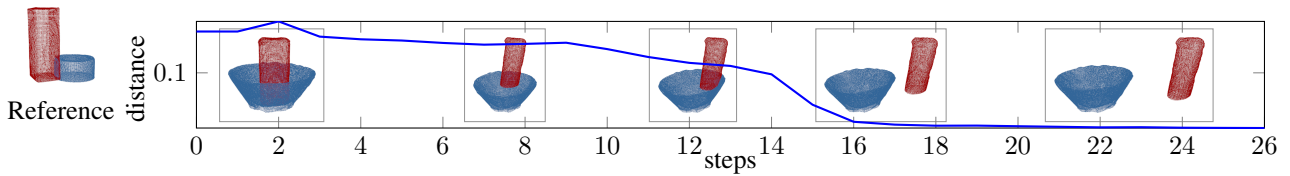
\subsection{Nearest Neighbor Retrieval}
We train the distance metric function on the Freiburg Spatial Relations dataset~\cite{mees2017metric}, which features 546 scenes each containing two out of 25 household objects. The dataset contains seven labeled relations, which we transform into a binary similarity matrix. As stated before, we train the metric using a triplet network. The network parameters $\mat{W}$ are optimized for $14,000$ iterations with a batch size of $100$ triplets. We sample the batches such that the provided class annotations of the scenes are uniformly distributed. Further, we apply data augmentation in a way that does not change the underlying ground truth spatial relation, i.e., we add a small amount of noise on the transformations of the objects and rotate the full scene around the z axis. Additionally, we apply dropout with a probability of $50\%$ on the fully-connected layer. For training we use stochastic gradient descent with a momentum of $0.9$ and warm restarts~\cite{loshchilov2017sgdr} with an initial learning rate of $0.001$ and a period length of 1500 steps, which is doubled after each restart.
We cross-validate the nearest neighbor performance of our learned metric on the 15 train/test splits of the Freiburg Spatial Relations Dataset provided with the dataset. As performance measure, we compute the mean 3-out-of-5 and 5-out-of-5 accuracy for nearest neighbor retrieval over the fifteen splits.
Table~\ref{tab:nnresults} shows that our method yields an accuracy of
$91.21\% \pm 2.78\%$, which is a relative improvement of $3.6\%$ compared to the GBLMNN approach by Mees~\etal~\cite{mees2017metric}. 
Our results show that the learned metric allows us to retrieve similar scenes from a continuous spectrum of relations in the learned space with high accuracy. Further, they suggest that the learned feature representation of the metric, captured in the last fully-connected layer of the network siblings (see Fig.~\ref{fig:Architecture}), is rich enough to be leveraged for gradient-based optimization in the next experiments.\\

\subsection{Generalizing Relations to Known Objects}
\label{results:generalizing}
Next, we quantitatively evaluate the capability of our approach to imitate spatial relations. For testing we randomly selected 13 scenes including 15 different objects such that every scene was similar to at most one other scene. 
We then considered all 156 combinations of these 13 scenes excluding 31 scenes that cannot be transformed into each other, e.g., a plate and a cup cannot be generalized to an inside relation. From the remaining 125 combinations, we used one scene as a reference to generalize the other scene, as qualitatively depicted in Fig.~\ref{fig:Generalization}. To compute the generalization, we used the Adam Optimizer with a learning rate of $0.1$ to minimize the metric distance by optimizing $\mathbf{t}_m, \mathbf{t}_n$ and $\mathbf{q}_m, \mathbf{q}_n$ of the test scene. Overall, 70 of the imitations successfully generalized the reference scene. 41 of these imitated scenes were physically infeasible scenes, e.g., containing objects placed on their edges. However, we do not account for scene stability or feasibility in this work and therefore consider them successful. 55 of the generalizations converged to a non-optimal solution.
Fig.~\ref{fig:Generalization} qualitatively depicts exemplary results for each. Among successful generalizations the figure shows a bowl that is correctly optimized into a tray and several inclined object relations. Note that both object transformations are optimized in a continuous manner, without specifying 
any semantic knowledge, see Fig.~\ref{fig:Optimization}.
Despite the fact that we do not provide knowledge about physical concepts such as collisions during the training process and despite the fact that we approximate the 3D world using our 2D projection technique, our approach is able to leverage the learned metric to generalize relations in the 3D domain. 

In addition, we conducted a real-world experiment with a Kinect2 camera, where we demonstrated reference scenes containing common spatial relations, see Fig.~\ref{fig:SmallCover}. To retrieve the point clouds and the poses of the reference objects we used the Simtrack framework~\cite{pauwels2015simtrack}. In this experiment we demonstrate the capability of our method to successfully generalize spatial relations in real-time. In a further experiment, we employed our framework on a real PR2 robot and used the robot to manipulate the objects of the test scene, using out-of-the-box motion planning.
A video of these experiments is available at~\url{http://spatialrelations.cs.uni-freiburg.de}.
\\

\subsection{Generalizing Relations to Unknown Objects}
\newcommand{\genstan}[4] {
    \node (ref) at (0,#1) [minimum width=1cm,minimum height=1cm] {
            \includegraphics[height=1cm, keepaspectratio, interpolate=false]{figures/generalization_stanford_objs/#2}};

    \node (init) [minimum width=1cm,minimum height=1cm, right=of ref] {
            \includegraphics[height=1cm, keepaspectratio, interpolate=false]{figures/generalization_stanford_objs/#3}};

    \node (gen) [minimum width=1cm,minimum height=1cm, right of =init, xshift=1.8cm] {
            \includegraphics[height=1cm, keepaspectratio, interpolate=false]{figures/generalization_stanford_objs/#4}};
    \draw[->, >=latex] (init) -- (gen);

}

\begin{figure}[!t]
\begin{center}
    \scalebox{0.85}{
    \begin{tikzpicture}[every node/.style = {anchor= north west}]
        \genstan{0}{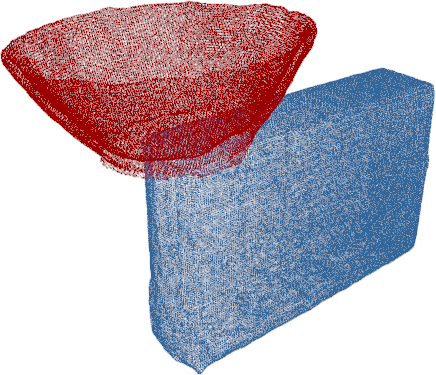}{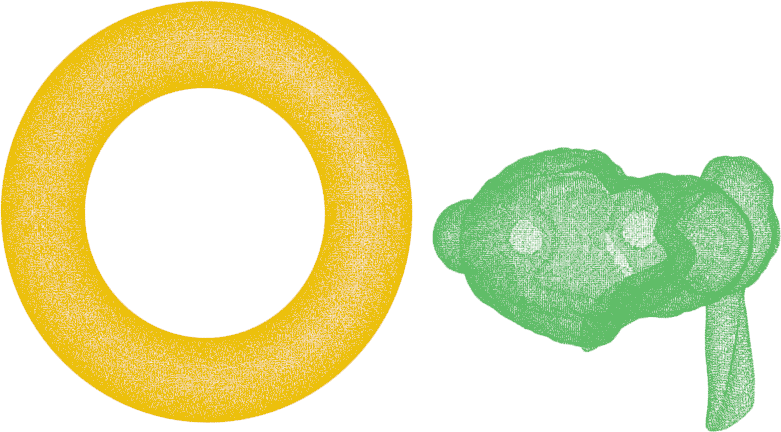}{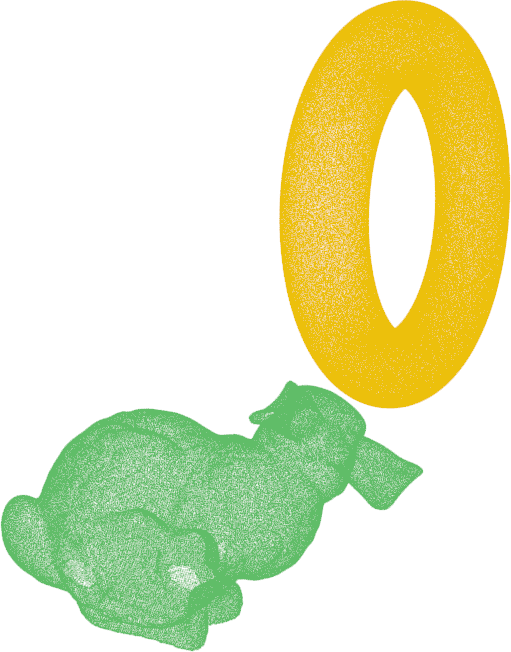}
        \node (reftext) [above=10pt of ref] {Reference};
        \node (inittext) [above=10pt of init] {Initial};
        \node (gentext) [above=10pt of gen] {Generalized};
        \draw (-0.3, 0.1) -- (6.3, 0.1);
        \genstan{-1.7}{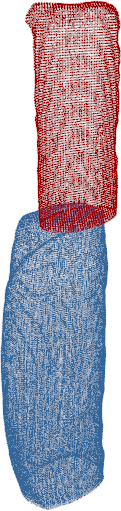}{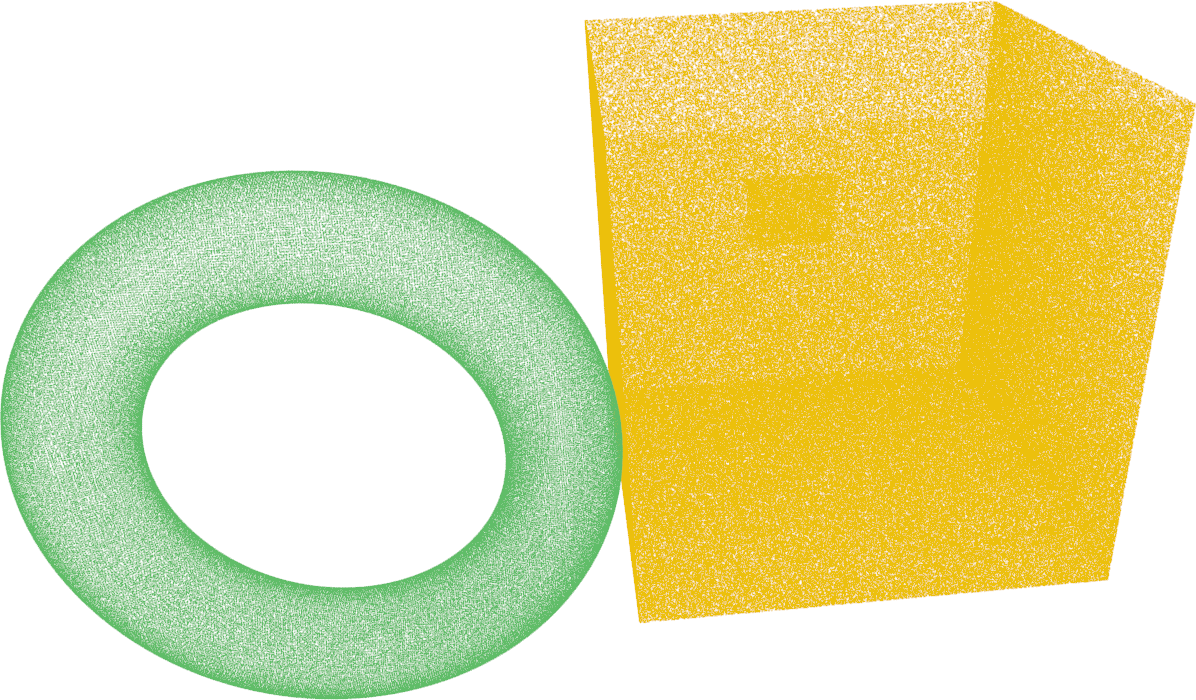}{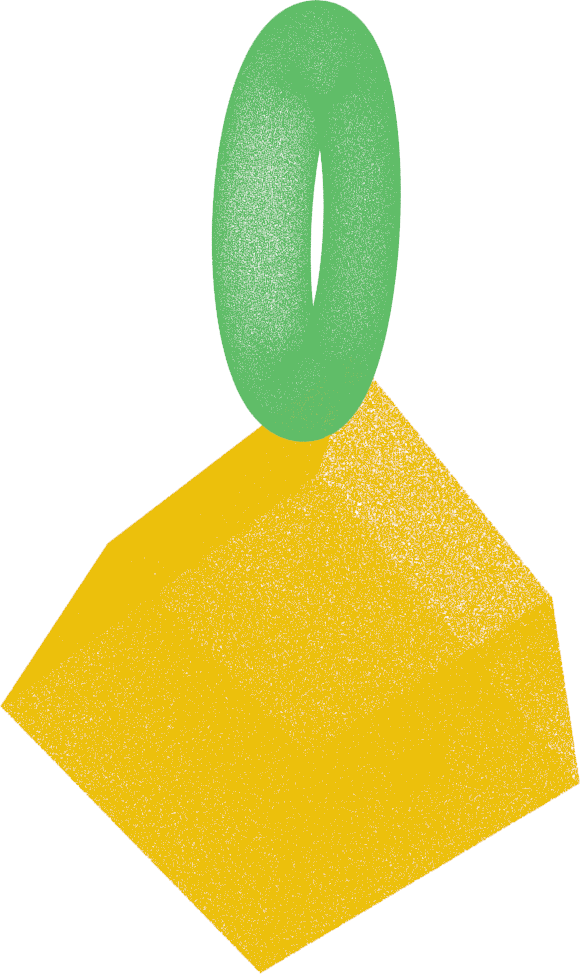}
        \genstan{-3.5}{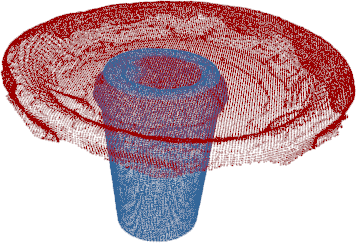}{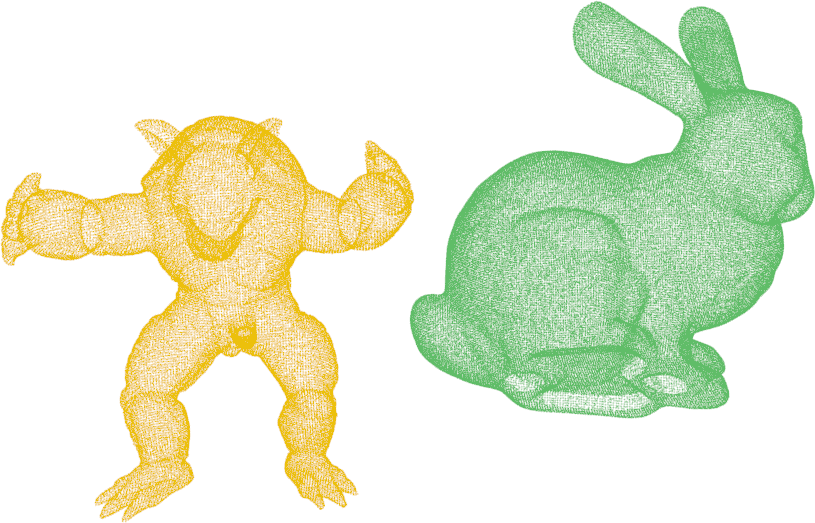}{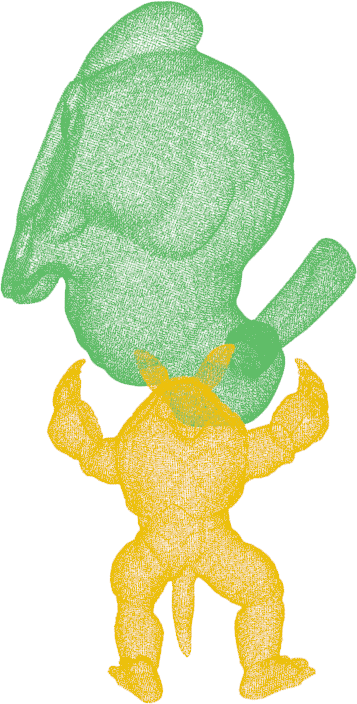}
        \genstan{-5.3}{0_shampoo2_sweetener_881.png}{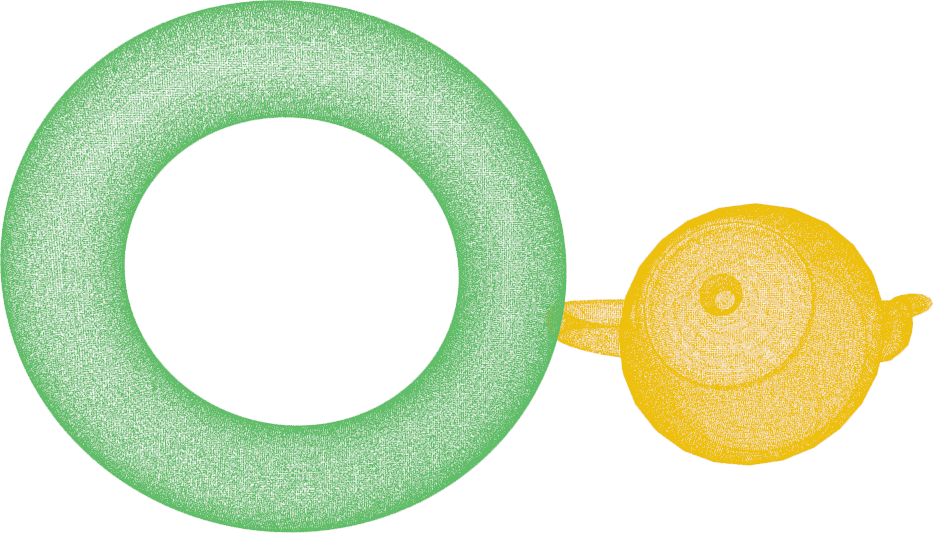}{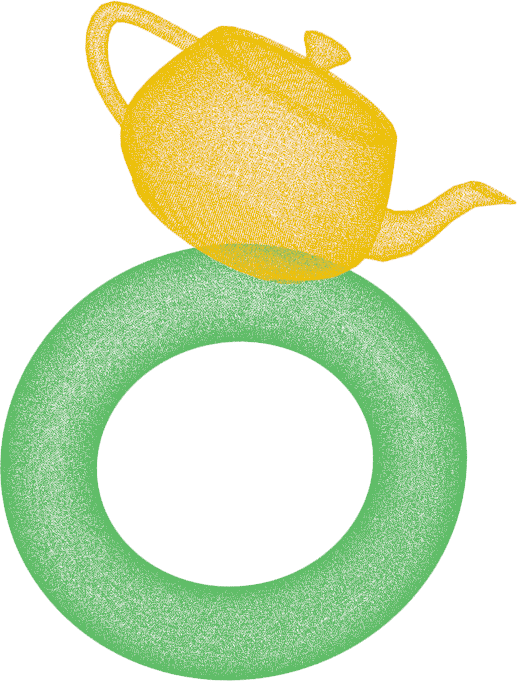}
        \genstan{-7}{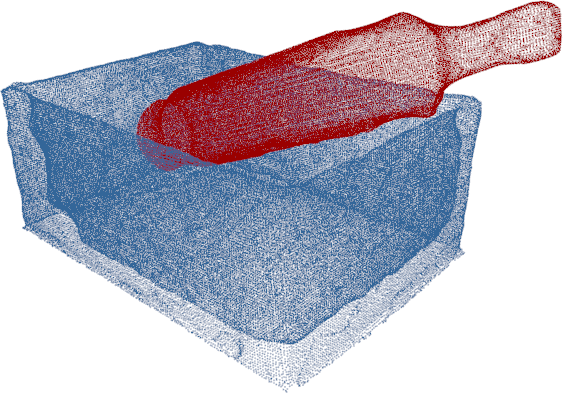}{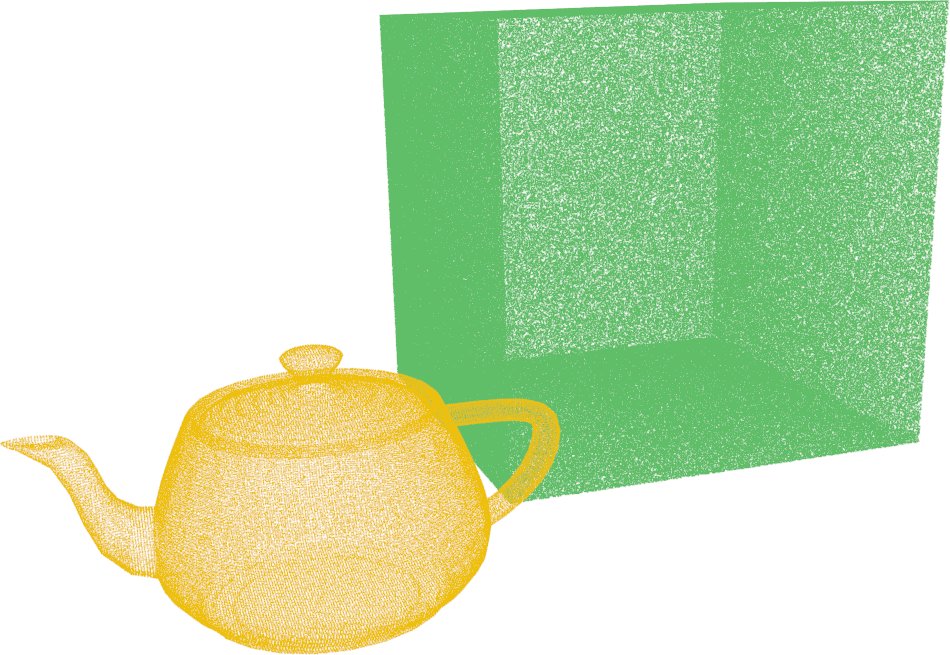}{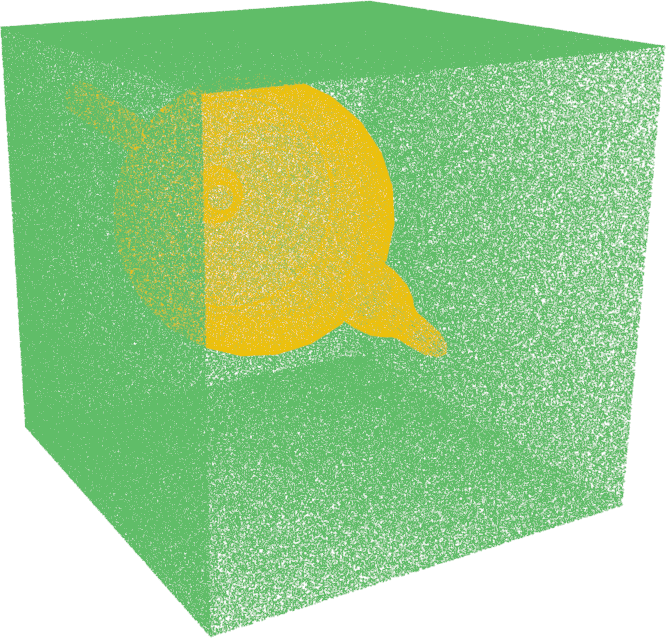}
    \end{tikzpicture}
    }
\end{center}

 \caption{Examples of four successfully and one unsuccessfully generalized relations to objects unseen during training. 
          Despite the complex, unknown shapes, the optimized scenes capture the semantic of the relation. The last example is counted as unsuccessful 
          because the opening of the cube should point upwards.}
\label{fig:GeneralizationUnkown}
\end{figure}

To evaluate how well our approach handles previously unseen objects, we chose five common 3D models such as the Utah tea pot and the Stanford bunny, as shown in Fig.~\ref{fig:GeneralizationUnkown}. We emphasize that the shapes of these objects differ notably from the objects of the Freiburg Spatial Relations dataset used during training.
As reference scenes we used a subset of the 13 previously used scenes. As test scenes we used all two-permutations without replacement of the five new objects, sampled next to each other. We then considered all 160 transformable combinations of training and test scenes, excluding the test object combinations that cannot form an inside relation. Our approach was able to successfully generalize 68 scenes and failed on 92 scenes. As expected, this is a more challenging task compared to the previous experiment since none of the test object shapes were used in training. Additionally, the dataset used for training covers only a small subset of the space of possible object arrangements, posing a challenge on dealing with intermediate relations the approach encounters during optimization. Nonetheless, our approach is able to imitate the semantics of a given spatial relation with considerably different objects and generalize them without the need for prior knowledge.

%
\section{CONCLUSIONS}
\label{sec:conclusion}
%
In this paper, we presented a novel approach to learning the similarity between pairwise spatial relations in 3D space and to imitate arbitrary relations between objects. Our approach learns a metric that allows reasoning over a continuous spectrum of such relations. In this way, our work goes beyond the state of the art in that we do not require learning a model for each new spatial relation. Furthermore, our work enables learning the metric and using it to generalize relations in an end-to-end manner and without requiring pre-defined expert features. For this, we introduced a novel approach for backpropagating the gradient of the metric to optimize the 3D transformation parameters of two objects in a scene in order to imitate an arbitrary spatial relation between two other objects in a reference scene. We evaluated our approach extensively using both simulated and real-world data. Our results demonstrate the ability of our method to capture the similarities between relations and to generalize them to objects of arbitrary shapes and sizes, which is a crucial requirement for intelligent service robots to solve tasks in everyday environments. To incorporate physical constraints such as object collisions, it would be interesting to add differentiable physics in the future~\cite{degrave2016differentiable}.

%

\section*{ACKNOWLEDGMENT}
We thank Marc Toussaint for the inspiring discussions in the early stages of this work,
Tobias Springenberg for fruitful discussion and feedback on leveraging 2D projections
and Oier Mees for providing the data set.


\bibliographystyle{ieeetr}
\bibliography{spatial_relations}  

\end{document}